\pdfoutput=1
\documentclass[10pt,twocolumn,letterpaper]{article}

\usepackage{cvpr}
\usepackage{times}
\usepackage{subcaption}
\usepackage{epsfig}
\usepackage{graphicx}
\usepackage{amsmath}
\usepackage{amssymb}
\usepackage{float}
\usepackage{authblk}

\usepackage{adjustbox}      
\usepackage{tikz}
\usepackage{tikzscale}%
\usetikzlibrary{arrows,shapes,chains,matrix,positioning}
\usetikzlibrary{scopes,decorations,shadows,backgrounds,fit}
\usetikzlibrary{decorations.pathreplacing,calc,3d}

\usepackage{soul} 

\usepackage{verbatim} 

\makeatletter
\renewcommand\AB@affilsepx{, \protect\Affilfont}
\makeatother

\renewcommand*{\Authsep}{, }
\renewcommand*{\Authand}{, }
\renewcommand*{\Authands}{, }
\renewcommand*{\Affilfont}{\normalsize}

\makeatletter
\pgfkeys{/pgf/.cd,
  parallelepiped offset x/.initial=2mm,
  parallelepiped offset y/.initial=2mm
}
\pgfdeclareshape{parallelepiped}
{
  \inheritsavedanchors[from=rectangle] 
  \inheritanchorborder[from=rectangle]
  \inheritanchor[from=rectangle]{north}
  \inheritanchor[from=rectangle]{north west}
  \inheritanchor[from=rectangle]{north east}
  \inheritanchor[from=rectangle]{center}
  \inheritanchor[from=rectangle]{west}
  \inheritanchor[from=rectangle]{east}
  \inheritanchor[from=rectangle]{mid}
  \inheritanchor[from=rectangle]{mid west}
  \inheritanchor[from=rectangle]{mid east}
  \inheritanchor[from=rectangle]{base}
  \inheritanchor[from=rectangle]{base west}
  \inheritanchor[from=rectangle]{base east}
  \inheritanchor[from=rectangle]{south}
  \inheritanchor[from=rectangle]{south west}
  \inheritanchor[from=rectangle]{south east}
  \backgroundpath{
    \southwest \pgf@xa=\pgf@x \pgf@ya=\pgf@y
    \northeast \pgf@xb=\pgf@x \pgf@yb=\pgf@y
    \pgfmathsetlength\pgfutil@tempdima{\pgfkeysvalueof{/pgf/parallelepiped offset x}}
    \pgfmathsetlength\pgfutil@tempdimb{\pgfkeysvalueof{/pgf/parallelepiped offset y}}
    \def\ppd@offset{\pgfpoint{\pgfutil@tempdima}{\pgfutil@tempdimb}}
    \pgfpathmoveto{\pgfqpoint{\pgf@xa}{\pgf@ya}}
    \pgfpathlineto{\pgfqpoint{\pgf@xb}{\pgf@ya}}
    \pgfpathlineto{\pgfqpoint{\pgf@xb}{\pgf@yb}}
    \pgfpathlineto{\pgfqpoint{\pgf@xa}{\pgf@yb}}
    \pgfpathclose
    \pgfpathmoveto{\pgfqpoint{\pgf@xb}{\pgf@ya}}
    \pgfpathlineto{\pgfpointadd{\pgfpoint{\pgf@xb}{\pgf@ya}}{\ppd@offset}}
    \pgfpathlineto{\pgfpointadd{\pgfpoint{\pgf@xb}{\pgf@yb}}{\ppd@offset}}
    \pgfpathlineto{\pgfpointadd{\pgfpoint{\pgf@xa}{\pgf@yb}}{\ppd@offset}}
    \pgfpathlineto{\pgfqpoint{\pgf@xa}{\pgf@yb}}
    \pgfpathmoveto{\pgfqpoint{\pgf@xb}{\pgf@yb}}
    \pgfpathlineto{\pgfpointadd{\pgfpoint{\pgf@xb}{\pgf@yb}}{\ppd@offset}}
  }
}
\makeatother

\usepackage[pagebackref=true,breaklinks=true,letterpaper=true,colorlinks,bookmarks=false]{hyperref}

\cvprfinalcopy 

\def\cvprPaperID{1716} 

\begin{document}

\title{Visual Compiler:\\Synthesizing a Scene-Specific Pedestrian Detector and Pose Estimator}


\author[1]{Namhoon Lee}
\author[2]{Xinshuo Weng}
\author[3]{Vishnu Naresh Boddeti}
\author[2]{Yu Zhang}
\author[4]{Fares Beainy}
\author[2]{Kris Kitani}
\author[2]{Takeo Kanade}
\affil[1]{University of Oxford}
\affil[2]{Carnegie Mellon University}
\affil[3]{Michigan State University}
\affil[4]{Volvo Construction}

\maketitle

\begin{abstract}
We introduce the concept of a Visual Compiler that generates a scene specific pedestrian detector and pose estimator without any pedestrian observations. Given a single image and auxiliary scene information in the form of camera parameters and geometric layout of the scene, the Visual Compiler first infers geometrically and photometrically accurate images of humans in that scene through the use of computer graphics rendering. Using these renders we learn a scene-and-region specific spatially-varying fully convolutional neural network, for simultaneous detection, pose estimation and segmentation of pedestrians. We demonstrate that when real human annotated data is scarce or non-existent, our data generation strategy can provide an excellent solution for bootstrapping human detection and pose estimation. Experimental results show that our approach outperforms off-the-shelf state-of-the-art pedestrian detectors and pose estimators that are trained on real data.
\end{abstract}

\section{Introduction}

Over the past decade, computer vision has seen great strides across a wide array of tasks including object recognition and detection \cite{krizhevsky2012imagenet}, semantic segmentation \cite{long2015fully}, image captioning \cite{karpathy2015deep}, face recognition \cite{taigman2014deepface} and many more. The success of these models depends heavily on the availability of computational resources and a key ingredient for learning such complex models -- large amounts of human annotated data. However, in many scenarios, unfortunately, it still remains that human labeled data is scarce or worse yet, simply unavailable.

\begin{figure}[t]
    \centering
    \includegraphics[width=1.0\linewidth]{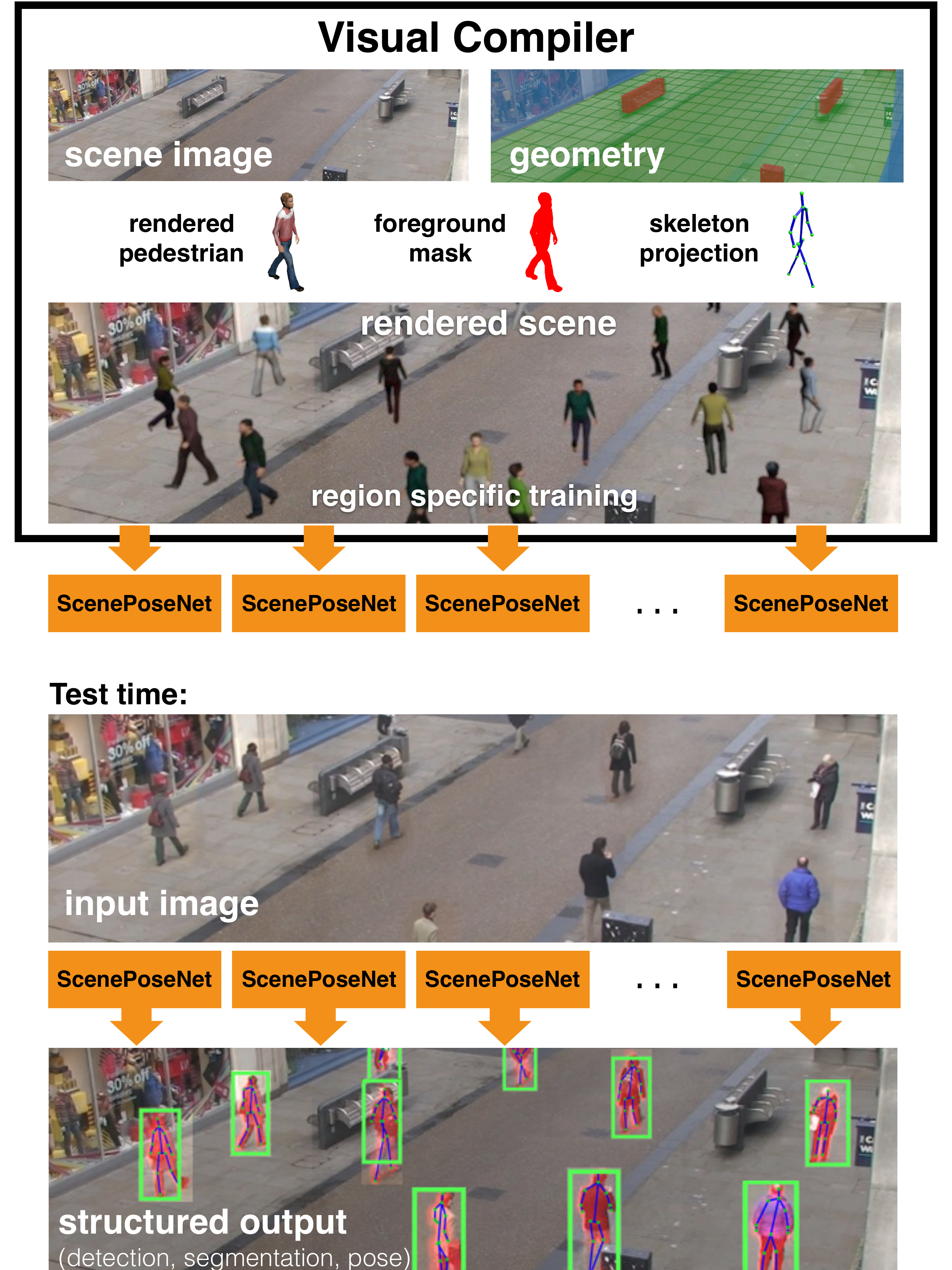}
    \caption{Overview: Given a single image of a scene, camera parameters and coarse scene geometry as input, the \emph{Visual Compiler} synthesizes physically grounded and geometrically accurate renders of pedestrians. Region specific pose networks are trained on this synthetic data. At test time, our model takes a single image and outputs pedestrian detections, segmentation mask and body pose estimates.}
    \label{fig:overview}
\end{figure}

In this paper we consider one such scenario where we must bootstrap a pedestrian detector and pose estimator for a specific surveillance environment without access to any real pedestrian data, either labeled or unlabeled. Consider the case where a new camera surveillance system has been installed, perhaps inside a building or perched above a street. We would like to use this camera to localize pedestrians and estimate their pose for high-level activity analysis. A straightforward solution would be to use an existing generic pedestrian detection and pose estimation system. Most of these generic systems however, are trained on a data distribution that is potentially quite different from the scene under consideration and may result in a very low accuracy system. Although it would be possible to adapt generic models to the new environment by incrementally labeling test examples, it would still require manual human intervention. In contrast, in this paper we would like to completely automate this process and learn a pedestrian detector and pose estimator without leveraging any real data.

We introduce the \emph{visual compiler} to address this problem. In the same way a computer language compiler takes directives in one language to generate a program that runs on the operating system (OS), our visual compiler takes the description of a scene (visual language) and transforms it into a pedestrian detector and pose estimator (visual program) appropriate for that scene (visual OS). The input to the \emph{visual compiler} is a scene description in the form of a single image, camera parameters and coarse geometric layout of the scene. Using this scene description, the \emph{visual compiler} uses a computer graphics rendering engine along with a customizable database of virtual human models to generate an endless number of data samples, thereby overcoming data scarcity through synthesis. In contrast to previous work that use 3D models for data synthesis, the key feature of our \emph{visual compiler} is that it generates physically grounded and geometrically accurate renders of humans in the scene under consideration.

Using geometrically consistent synthesis of humans presents us with many advantages that can compensate for the lack of real training data: (1) We can maximize the physical geometric information in the scene in terms of the appearance of humans in the scene, the static objects in the scene causing occlusions, resolution and quality of human appearance captured by the camera system, distortions caused by camera optics and partial people at the edges of the camera frame. This geometric information can be incorporated into the data synthesis pipeline to generate realistic renders of virtual humans. (2) We can potentially synthesize an unlimited amount of pedestrian samples spanning a wide range of appearance variations (\eg., clothing, height, weight, gender, ethnicity) on demand. (3) We can simulate human appearance at literally all potential locations in the scene that humans can exist. Additionally we can precisely control the pose, orientation and 3D location of the simulated pedestrian in the scene. (4) We can automatically obtain annotations for detection, body part locations and segmentation masks. The annotations obtained this way are noiseless and precise while human-labeled data is often noisy and error prone.

Using the high quality synthetic images of pedestrian appearance in the scene, the visual compiler learns a scene-and-region specific spatially-varying fully convolutional neural network, dubbed ScenePoseNet, for simultaneous detection, pose estimation and segmentation of pedestrians. Traditionally synthetic data has often been used in conjunction with real data while training, either for learning models from scratch or for fine tuning an existing model. In contrast, ScenePoseNet is trained purely on synthetic data from scratch. Surprisingly, our method outperforms competitive alternatives that are trained on real data, when evaluated not only on synthetic images but on real data as well, contradicting conventional wisdom that models purely on synthetic data is not sufficient for high accuracy.

\section{Related Work}

The use of synthetic models has been explored for a variety of computer vision tasks, typically in the context of data augmentation or domain adaptation for object classification. Aubry \emph{et al.} \cite{aubry2014seeing} posed object detection as a 2D-3D alignment problem and learned exemplar classifiers from 3D models to align and retrieve the models that best matches the viewpoint of 2D objects in images. Vasquez \emph{et al.} \cite{vazquez2014virtual} combined synthetic pedestrian data with real pedestrian data to generate robust real world detectors. Pishchulin \emph{et al.} \cite{pishchulin2012articulated} generated pedestrian samples with realistic appearance and backgrounds while modifying body shape and pose using 3D models to augment their real training data for pose estimation. These techniques demonstrated that the performance of visual classifiers can be improved by augmenting real data with a large amount of synthetic data. We emphasize here that we operate in a different regime where \textit{no real data is available} for augmentation or adaptation.

Visual analysis tasks can also be trained using only synthetic data. Recently Su \emph{et al.} \cite{su2015render} proposed to use a large collection of 3D models for viewpoint estimation in images. Fischer \emph{et al.} \cite{fischer2015flownet} used rendered data of flying chairs for supervised optical flow prediction. However, in these tasks the rendering is done without considering any scene information which results in physically implausible synthetic images (\emph{e.g.}, floating cars). Shotton \emph{et al.} \cite{shotton2013efficient} leveraged prior knowledge that the camera will be roughly fronto-parallel to the user to generate a variety of synthetic depth maps to train a human pose estimator. Hattori \emph{et al.} \cite{hattori2015learning} used prior information about the scene to learn scene-specific pedestrian detectors purely from synthetic data. The work showed that leveraging prior camera and scene knowledge in the synthetic data generation pipeline can help to ensure a tighter coupling between people observed in the training and testing data distributions. Our approach builds on the work of \cite{hattori2015learning} but extends to a far more challenging task, \ie, simultaneous articulated human pose estimation and body segmentation in addition to detection. Furthermore, our proposed model is based on a deep convolutional neural network that is trainable end-to-end instead of using a support vector machine on top of hand-crafted features.

There is a large body of work for pedestrian detection and human pose estimation. A complete treatment of this vast literature is beyond the scope of this paper. We instead provide a brief overview of the main techniques and focus on the most relevant state-of-the-art methods. Research on pedestrian detection is largely focused on designing better feature representations and part-based architectures. Carefully designed features \cite{dalal2005histograms,dollar2009integral,zhang2015filtered} that are computationally efficient have been the focus of much of the last decade. In contrast, modern day methods for pedestrian detection are based on carefully designed deep network architectures for feature learning \cite{tian2015deep,cai2015learning}. Architecturally, deformable part based methods \cite{girshick2011object,ouyang2013joint} have been the dominant method for detecting pedestrians. More recently, it has been shown that general object detection frameworks \cite{ren2015faster, wei2016ssd, girshick2015fast} can also achieve competitive pedestrian detection performance.


Interestingly, techniques for human pose estimation have been developed independently from human detection, where it is often assumed that a rough or the ground-truth location of target is available prior to pose estimation. Techniques for human pose estimation can be largely categorized into deformable parts based models \cite{felzenszwalb2005pictorial,yang2013articulated,pishchulin2013strong,wei2016deformablepose}, deep convolutional networks that regress from the image to the keypoint locations \cite{toshev2014deeppose,IEF2015human} and methods that regress from the image to the ideal localization heat-maps \cite{ramakrishna2014pose,wei2016cpm,newell2016stacked} of body parts. Toshev \emph{et al.} \cite{toshev2014deeppose} introduced one of the earliest deep learning based approaches for pose estimation, learning a regression function from the image to the part coordinates. Carreira \emph{et al.} \cite{IEF2015human} introduced a similar approach that iteratively refines the prediction of part locations. Current state-of-the-art approaches for human pose estimation, Convolutional Pose Machines (CPM) \cite{wei2016cpm} and Stacked Hourglass Networks \cite{newell2016stacked}, directly regress part localization heat maps from the input image. These approaches, 1) assume that humans have been detected, at least coarsely, and 2) are trained on real annotated images spanning a range of human pose and appearance.

Our \emph{Visual Compiler}, learns a scene-and-region-specific model that integrates (via heat map regression) pedestrian detection, pose estimation and segmentation into a single fully convolutional neural network. And unlike existing approaches, our model is trained purely on \textbf{synthetically rendered} pedestrians and evaluated on \textbf{real} pedestrian images. By leveraging geometrically accurate renderings of humans in the scene, our approach is able to bridge the gap in appearance between real and synthetic humans and outperforms generic state-of-the-art approaches for human detection and pose estimation for a given scene.

\section{Visual Compiler}

Figure \ref{fig:overview} gives a pictorial illustration of the inner workings of our visual compiler to generate a scene-specific human detection and pose estimation system given a scene description. We consider a setting where the following information is known \emph{a priori} (however, automated ways of obtaining this information exist): (1) the parameters of the camera, both intrinsic and extrinsic, (2) coarse physical geometric layout of the scene in terms of the regions of the scene where people could potentially exist (walking, sitting, standing) and regions where they could potentially be occluded (obstacles) or cannot physically exist (walls), and optionally (3) priors on the pose and orientation of pedestrians at various regions of the scene. Along with a single image, this scene description serves as the input to the compiler to synthesize physically grounded and geometrically accurate humans in the valid regions of the scene. The compiler then learns an ensemble of region-specific models for simultaneous detection, pose estimation and segmentation of humans. During inference, each of these region-specific models are run in parallel on their corresponding regions.

\subsection{Data Synthesis from Scene Description}

High quality ground truth annotations are required to train pedestrian detection and pose estimation systems. Obtaining these labels from real data usually requires a costly and noisy process of manual human labeling, a process that does not scale very well to a large number of scenes. Instead, the visual compiler uses the scene description to simulate probable pedestrian appearance appropriate for each region of the scene.

Given the scene description, the compiler first generates a planar 3D model of the scene, \emph{i.e.}, fits a planar ground plane, planar walls and cuboids to encompass the obstacles. The camera parameters can then be used to account for camera lens characteristics (\emph{e.g.}, perspective distortion in wide-angle cameras) and the scene viewpoint for rendering geometrically accurate humans. Autodesk 3DS Max is used as the scene modeling and rendering engine. The rendering pipeline can precisely control the following variations in human appearance: gender, height, width, orientation and pose in addition to rendering human appearance at every ``valid pedestrian location" of the scene. The virtual human database consists of 139 different models spanning gender, clothing color and ethnicity. The models used for this work only have skin tight clothing but have a continuous range of walking configurations from standing to running. The compiler uniformly samples body orientations from $0^{\circ} \sim 360^{\circ}$ but can also be guided by any prior information if available.

To generate ground truth labels for the people in the rendered images we first associate attributes to each 3D virtual model with the following labels: segmentation mask, 3D locations of 27 parts and the location of the center of the person for detection. The 2D labels for training can then be automatically extracted from the 3D annotations and the camera projection parameters. This process allows us to generate consistent noise free labels, unlike human annotations, at scale across all rendered images. Furthermore, we can also uniformly span all the variations in appearance, orientation, pose or location unlike real data that follows a long-tailed distribution.

\subsection{Learning the Network from Synthetic Data}

Using the scene-specific data generated above, the visual compiler now generates a visual program, in the form of deep neural network, trained to operate according to the specifications of the scene description.

The visual program generated by the visual compiler is designed to jointly accomplish the following tasks: localization of pedestrians, localizing the landmarks that define their pose and segment the pixels that define them. To predict the pedestrian location, pose and segmentation mask the network has to model the full appearance of the pedestrian, the local appearance of the landmarks and a prior on the valid spatial configuration of these parts. The network design aims to encapsulate these desiderata. To capture appearance, both the full pedestrian and the local landmark appearance, learning is posed as a regression problem mapping the RGB input into a heatmap for accurate localization of the pedestrian, local landmarks and the segmentation mask. The prior on the spatial relationships between the part locations is implicitly learned through a spatial belief module that accounts for the correlations between the heatmaps of the full pedestrian, local landmarks and the segmentation masks. We call this specific instantiation of a visual program as ScenePoseNet.

Human pose estimation systems typically treat detection and pose estimation as independent and sequential tasks, with detection followed by pose estimation. These systems either expect ground truth human detections or at least expect a coarse detection using an off-the-shelf detector. However, the tasks of detection and part localization are highly interdependent processes. Detection can greatly affect the pose estimation process and accurate localization of parts serves to enhance the belief of human presence at a corresponding location. Accordingly, the ScenePoseNet model couples these tasks to improve the efficacy of both pedestrian detection and pose estimation. The main idea behind our ScenePoseNet architecture is to (1) regress part localization beliefs from the image features and (2) learn the interactions between these confidence maps.

\subsection{Basic blocks}

\tikzstyle{mybox} = [draw=red, fill=blue!20, very thick,rectangle, rounded corners, inner sep=10pt, inner ysep=20pt]
\tikzstyle{fancytitle} =[fill=red, text=white]

\tikzstyle{concat} = [rectangle, draw, fill=yellow!20, text width=1.0em, minimum height=1.0em, text centered, drop shadow]
\tikzstyle{conv} = [rectangle, draw, fill=blue!20, text width=2.0em, text centered, minimum height=6em, drop shadow]
\tikzstyle{batchnorm} = [rectangle, draw, fill=green!20, text width=2.0em, text centered, minimum height=6em, drop shadow]
\tikzstyle{relu} = [rectangle, draw, fill=red!20, text width=2.0em, text centered, minimum height=6em, drop shadow]
\tikzstyle{circ} = [draw,circle,fill=teal!20,node distance=2cm, drop shadow]

\begin{figure}[t]
    \begin{subfigure}{\linewidth}
    \begin{adjustbox}{width=\textwidth}
    \begin{tikzpicture}[node distance=1.5cm, auto,>=latex', thick]
    \node (a) {};
    \node[right of=a] (b) {};
    \node[batchnorm, right of=b] (c) {\rotatebox{90}{BatchNorm}};
    \node[relu, right of=c] (d) {\rotatebox{90}{ReLU}};
    \node[conv, right of=d] (e) {\rotatebox{90}{3 x 3 Conv}};
    \node[batchnorm, right of=e] (f) {\rotatebox{90}{BatchNorm}};
    \node[relu, right of=f] (g) {\rotatebox{90}{ReLU}};
    \node[conv, right of=g] (h) {\rotatebox{90}{3 x 3 Conv}};
    \node[circ, right of=h] (i) {+};
    \node[right of=i] (j) {};
    \node[above of=b] (b1) {};
    \node[above of=i] (i1) {};
    
    \draw[->] (a) -- (c);
    \draw[->] (c) -- (d);
    \draw[->] (d) -- (e);
    \draw[->] (e) -- (f);
    \draw[->] (f) -- (g);
    \draw[->] (g) -- (h);
    \draw[->] (h) -- (i);
    \draw[->] (b)+(0,-0.02) -- (b1.north) -- (i1.north) -- (i);
    \draw[->] (i) -- (j);
    \end{tikzpicture}
    \end{adjustbox}
    \caption{Residual Module}
    \end{subfigure}
    \begin{subfigure}{\linewidth}
    \begin{adjustbox}{width=\textwidth}
    \begin{tikzpicture}[node distance=1.5cm, auto,>=latex', thick]
    \node (a) {};
    \node[right of=a] (b) {};
    \node[batchnorm, right of=b] (c) {\rotatebox{90}{BatchNorm}};
    \node[relu, right of=c] (d) {\rotatebox{90}{ReLU}};
    \node[conv, right of=d] (e) {\rotatebox{90}{17 x 17 Conv}};
    \node[batchnorm, right of=e] (f) {\rotatebox{90}{BatchNorm}};
    \node[relu, right of=f] (g) {\rotatebox{90}{ReLU}};
    \node[conv, right of=g] (h) {\rotatebox{90}{17 x 17 Conv}};
    \node[circ, right of=h] (i) {+};
    \node[concat, right of=i] (j) {+};
    \node[right of=j] (k) {};
    \node[above of=b] (b1) {};
    \node[above of=i] (i1) {};
    
    \node[below of=b, node distance=3cm] (b2) {};
    \node[batchnorm, below of=c, node distance=3cm] (c2) {\rotatebox{90}{BatchNorm}};
    \node[relu, below of=d, node distance=3cm] (d2) {\rotatebox{90}{ReLU}};
    \node[conv, below of=e, node distance=3cm] (e2) {\rotatebox{90}{1 x 1 Conv}};
    \node[batchnorm, below of=f, node distance=3cm] (f2) {\rotatebox{90}{BatchNorm}};
    \node[relu, below of=g, node distance=3cm] (g2) {\rotatebox{90}{ReLU}};
    \node[conv, below of=h, node distance=3cm] (h2) {\rotatebox{90}{1 x 1 Conv}};
    
    \draw[->] (a) -- (c);
    \draw[->] (c) -- (d);
    \draw[->] (d) -- (e);
    \draw[->] (e) -- (f);
    \draw[->] (f) -- (g);
    \draw[->] (g) -- (h);
    \draw[->] (h) -- (i);
    \draw[->] (b) -- (b1.north) -- (i1.north) -- (i);
    \draw[->] (i) -- (j);
    
    \draw[->] (b.north) |- (c2);
    \draw[->] (c2) -- (d2);
    \draw[->] (d2) -- (e2);
    \draw[->] (e2) -- (f2);
    \draw[->] (f2) -- (g2);
    \draw[->] (g2) -- (h2);
    \draw[->] (h2) -| (j);
    \draw[->] (j) -- (k);
    \end{tikzpicture}
    \end{adjustbox}
    \caption{Spatial Belief Module}
    \end{subfigure}
    \caption{The basic modules that comprise our ScenePoseNet architecture, (a) Residual module \cite{he2016identity}, (b) Spatial-Belief (SB) module. The spatial-belief unit aggregates ($\boxplus$ denotes the concatenation operation) the image features extracted from the convolutional network and the confidence maps of the full pedestrian, local landmarks and the segmentation mask from the output of the previous SB unit. The aggregated features now serve as the input to the next SB unit where the image features and confidence maps are jointly processed, thereby learning a prior on valid spatial relationships between the heat maps, and consequently body pose.}
    \label{fig:units}
\end{figure}
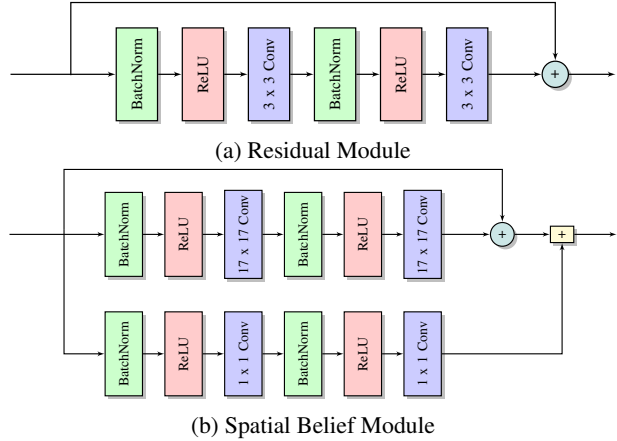


\tikzstyle{resunit} = [parallelepiped,draw=black,fill=brown!60,minimum width=0.1cm,minimum height=1.0cm]
\tikzstyle{cfunit} = [parallelepiped,draw=black,fill=violet!40,minimum width=0.1cm,minimum height=1.0cm]
\tikzstyle{conv} = [parallelepiped,draw=black,fill=blue!20,minimum width=0.1cm,minimum height=1.0cm]

\begin{figure*}
    \centering
    \begin{tikzpicture}[node distance=0.8cm, auto,>=latex', thick]
        
        \node[canvas is yx plane at z=0.0, transform shape] at (-0.4,0) {\scalebox{1}[-1]{\includegraphics[scale=0.1,angle=270]{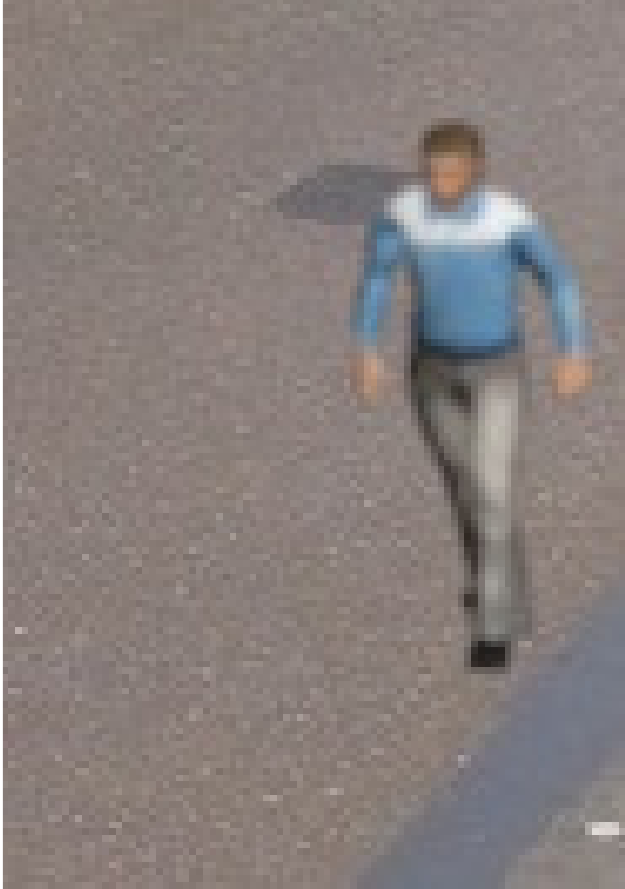}}};
        
        \node[] (start) {};
        \node[conv, right of=start, node distance=0.8cm] (1) {};
        
        \node[resunit,right of=1] (3) {};
        \node[resunit,right of=3] (4) {};
        \node[resunit,right of=4] (5) {};
        \node[resunit,right of=5] (6) {};
        
        \node[cfunit,right of=6] (7) {};
        \node[right of=7,node distance=0.4cm] (8) {};
        \node[cfunit,right of=7] (9) {};
        \node[right of=9,node distance=0.4cm] (10) {};
        \node[cfunit,right of=9] (11) {};
        \node[right of=11,node distance=0.4cm] (12) {};
        \node[conv, right of=11] (13) {};
        \node[conv, right of=13] (14) {};
        \node[concat,text width=1.0em, minimum height=1.0em,right of=14] (15) {+};
        \node[conv, right of=15] (16) {};
        \node[conv, right of=16] (17) {};
        
        \node[right of=17, node distance=0.8cm] (18) {};
        \node[canvas is yx plane at z=0.0, transform shape] at (12.0,0) {\scalebox{1}[-1]{\includegraphics[scale=0.1,angle=270]{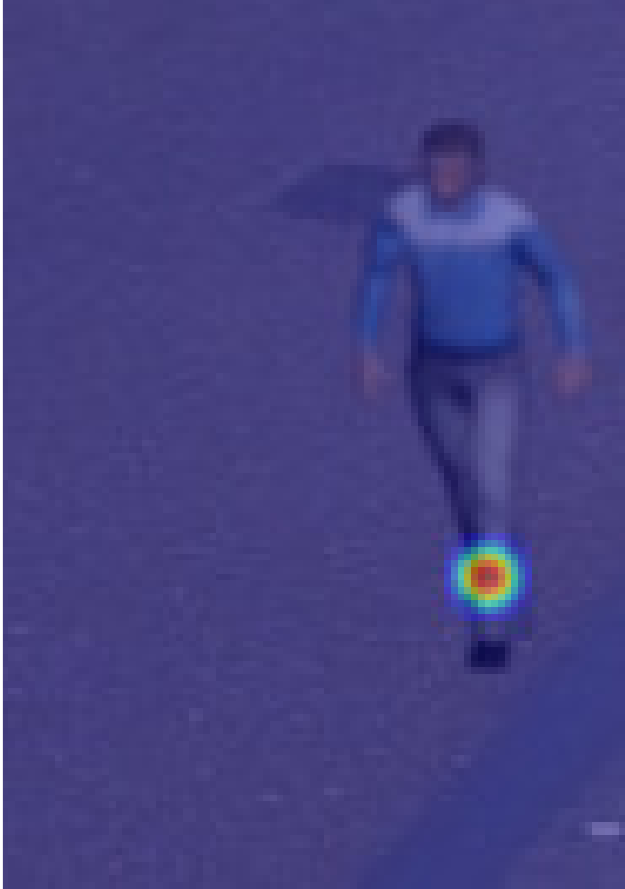}}};
        \node[canvas is yx plane at z=0.5, transform shape] at (12.0,0) {\scalebox{1}[-1]{\includegraphics[scale=0.1,angle=270]{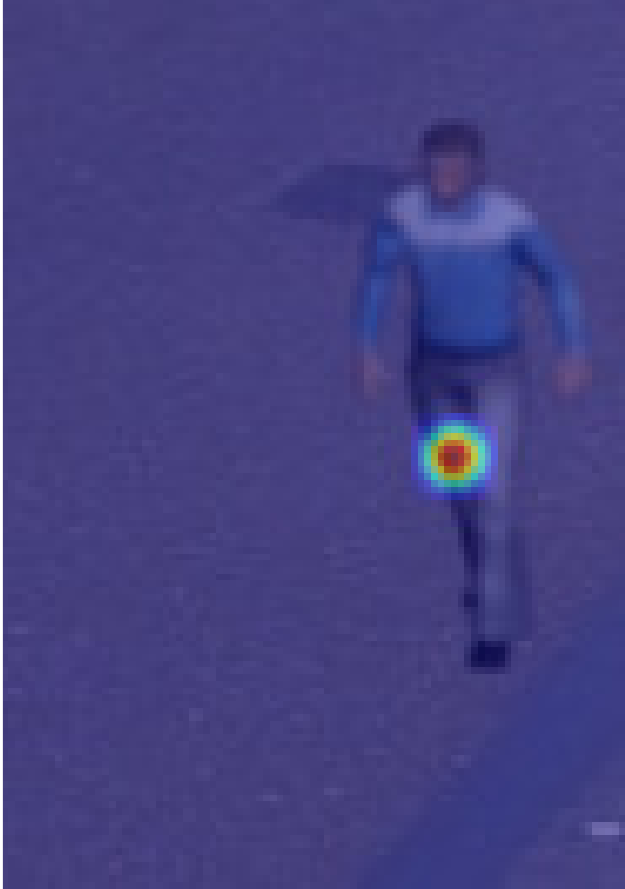}}};
        \node[canvas is yx plane at z=1.0, transform shape] at (12.0,0) {\scalebox{1}[-1]{\includegraphics[scale=0.1,angle=270]{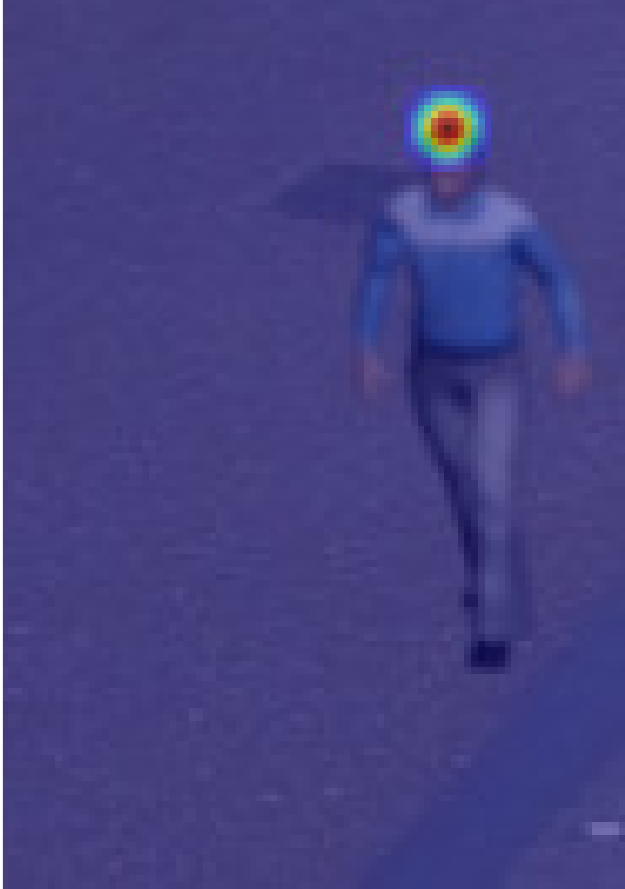}}}; 
        \node[canvas is yx plane at z=0.0, transform shape] at (13.0,0) {\scalebox{1}[-1]{\includegraphics[scale=0.1,angle=270]{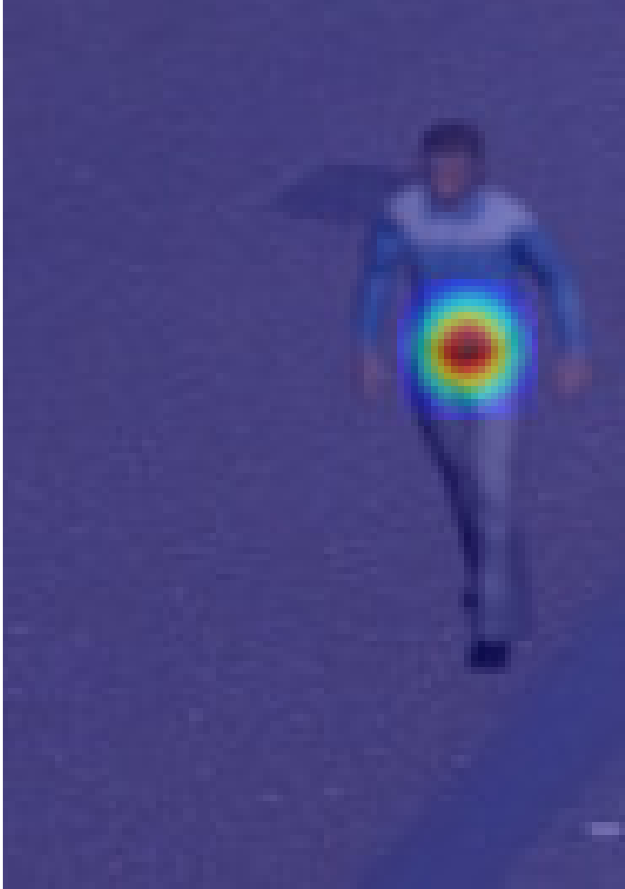}}};
        \node[canvas is yx plane at z=0.5, transform shape] at (12.8,0) {\scalebox{1}[-1]{\includegraphics[scale=0.1,angle=270]{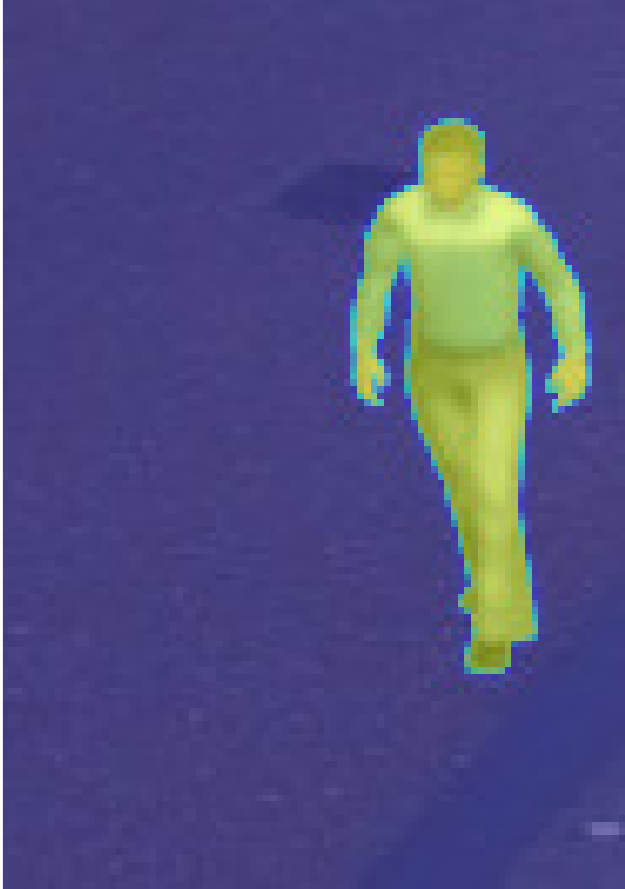}}};
        
        \node[below of=8,node distance=1.5cm] (a1) {};
        \node[below of=10,node distance=1.0cm] (b1) {};
        \node[below of=12,node distance=0.5cm] (c1) {};
        
        \node[below of=15,node distance=1.5cm] (a2) {};
        \node[below of=15,node distance=1.0cm] (b2) {};
        \node[below of=15,node distance=0.5cm] (c2) {};
        
        \draw[->] (start) -- (1);
        \draw[->] (1) -- (3);
        \draw[->] (3) -- (4);
        \draw[->] (4) -- (5);
        \draw[->] (5) -- (6);
        \draw[->] (6) -- (7);
        \draw[->] (7) -- (9);
        \draw[->] (9) -- (11);
        \draw[->] (11) -- (13);
        \draw[->] (13) -- (14);
        \draw[->] (14) -- (15);
        \draw[->] (15) -- (16);
        \draw[->] (16) -- (17);
        \draw[->] (17) -- (18);
        
        \draw[->] (8)+(0,0.02) -- (a1.south) -- (a2.south) -- (15);
        \draw[->] (10)+(0,0.02) -- (b1.south) -- (b2.south) -- (15);
        \draw[->] (12)+(0,0.02) -- (c1.south) -- (c2.south) -- (15);
        
    \end{tikzpicture}
    \caption{\textbf{ScenePoseNet:} An illustration of our network architecture. Our network is comprised of three basic units: \textcolor{blue}{convolutional} block, \textcolor{brown}{residual} block and \textcolor{violet}{spatial-belief} block. Our network uses information from multiple different spatial contextual regions via skip connections ($\boxplus$ denotes the concatenation operation). The input image is mapped to the ideal heat maps for part localization, pedestrian center and segmentation mask.}
    \label{fig:fullnet}
\end{figure*}
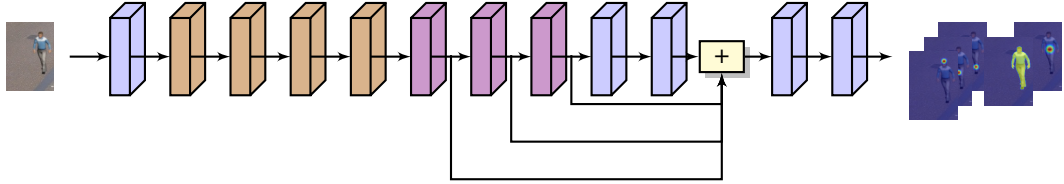

We use the following basic units to define our network: Residual Unit \cite{he2016identity} and Spatial-Belief Module. The residual unit was introduced to address the problem of vanishing gradients in training very deep convolutional networks. We adopt this basic unit for our network and also build upon it to define our Spatial-Belief (SB) module. As shown in Figure \ref{fig:units}(b) the SB module is purposed to (1) map the input features of the block to the part localization beliefs (heat-maps) and simultaneously (2) process the input features and part localization beliefs from the previous block. The image features and the part localization beliefs generated by this block are concatenated to form the input to the next block. Given an input $\mathbf{x}$ to SB module, the output $\mathbf{y}$ is given by,
\begin{align}
    \mathbf{y} &= (\mathbf{x} + f_{res}(\mathbf{x})) \boxplus f_{belief}(\mathbf{x}) \\
    &= (\mathbf{x} + \mathbf{r}) \boxplus \mathbf{b} 
\end{align}

\noindent where $\boxplus$ denotes the concatenation operation, $\mathbf{r} = f_{res}(\mathbf{x})$ is the operation through the non-identity branch of the residual unit and $\mathbf{b} = f_{belief}(\mathbf{x})$ denotes the mapping from the input $\mathbf{x}$ to the desired heat maps (human detection, part detection and segmentation mask) through a series of $1 \times 1$ convolutions. Our SB unit enables the network to consider part detection confidences with varying amounts of contextual information around the parts from different receptive fields. The part localization confidences $\mathbf{b}_i$ from the $i$-th SB unit propagates to the next ($i+1$)-th SB block and is processed through the non-identity path where the correlations between the heat-maps of the various parts are implicitly captured. This can be readily seen by applying the SB unit operation recursively,
\begin{align}
    \mathbf{x}_{i+1} = (\mathbf{x}_i + \mathbf{r}_i) \boxplus \mathbf{b}_i.
\end{align}
Both the identity shortcut and the $f_{res}()$ in each SB unit implicitly processes the beliefs from all previous SB units due to our concatenation operation. Furthermore, the detection confidence maps generated in each SB unit also consider part localization confidences at all previous SB units, each computed with different receptive fields. Therefore, the network takes advantage of detection confidence maps at multiple stages and through multiple receptive field sizes.

\subsection{ScenePoseNet}
Our complete detection, pose estimation and segmentation network architecture is illustrated in Figure~\ref{fig:fullnet}. Given an input image, ScenePoseNet jointly localizes pedestrians, localizes body parts and segments the pedestrians in the form of heat maps. The network is composed of fully convolutional layers to preserve spatial context while being computationally efficient. For precise localization and pose estimation of pedestrians we also use dense heat map prediction throughout the network preventing loss of information due to sub-sampling (pooling). The input image is passed through a convolutional layer with $5 \times 5$ filters, followed by four residual units with $3 \times 3$ filters following the design of residual networks for object recognition. This is followed by 3 SB units each with convolutional filters with large receptive fields, $17\times17$ to increase the receptive field of the network while still performing dense prediction. The SB units are followed by two $1 \times 1$ convolutional layers to map the image features to the heat maps. Finally, skip connections are used for fusing information from multiple different contextual regions, as it combines features from various scales of receptive fields (similar to~\cite{wei2016cpm}). The bounding box location for detection is inferred around the heatmaps of joints, center of body, and segmentation. 

The network is optimized to minimize the multi-task mean-squared-error loss $\mathcal{L}$ between the network prediction $\{\mathbf{o}_{det},\mathbf{o}_{pose},\mathbf{o}_{seg}\}=f_{conv}(\mathbf{b}_i\boxplus\cdots\boxplus\mathbf{b}_n)$ and the ideal heatmaps for pedestrian detection, part localization and segmentation mask, defined as follows,
\begin{equation}
    \mathcal{L} = \alpha\mathcal{L}_{det} + \beta\mathcal{L}_{pose} + \gamma\mathcal{L}_{seg} 
\end{equation}
\begin{align}
    \mathcal{L}_{det} & = \|\mathbf{o}_{det}-\mathbf{g}_{det}\|_2^2 \\ 
    \mathcal{L}_{pose} & = \frac{1}{n}\sum_{i=1}^n \|\mathbf{o}_{pose}-\mathbf{g}_{pose}\|_2^2 \\
    \mathcal{L}_{seg} & = \|\mathbf{o}_{seg}-\mathbf{g}_{seg}\|_2^2
\end{align}
where $\alpha$, $\beta$ and $\gamma$ are hyperparameters trading-off the different loss functions.

\section{Experiments and Analysis}

\begin{figure*}[ht]
    \centering
    \includegraphics[width=0.8\linewidth, trim={0 1.5cm 0 1.23cm},clip]{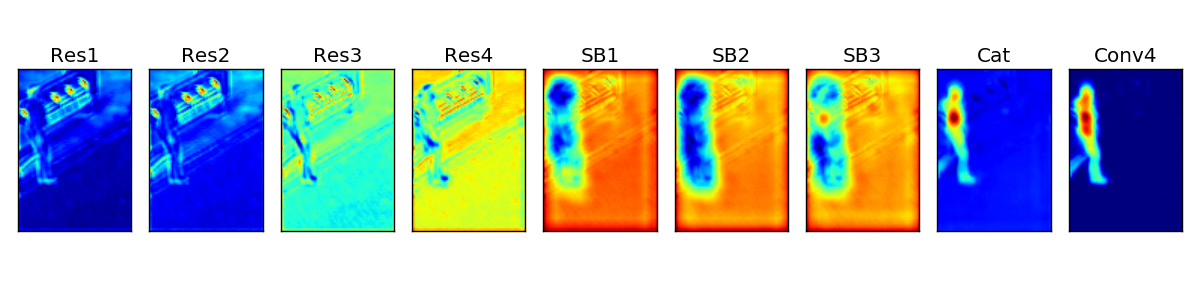}    \\
    \includegraphics[width=0.8\linewidth, trim={0 1.5cm 0 1.1cm},clip]{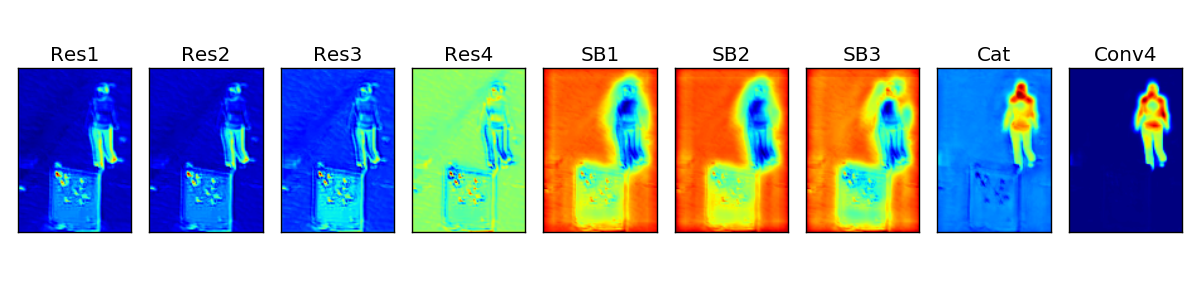}    \\
    \includegraphics[width=0.8\linewidth, trim={0 1.5cm 0 1.1cm},clip]{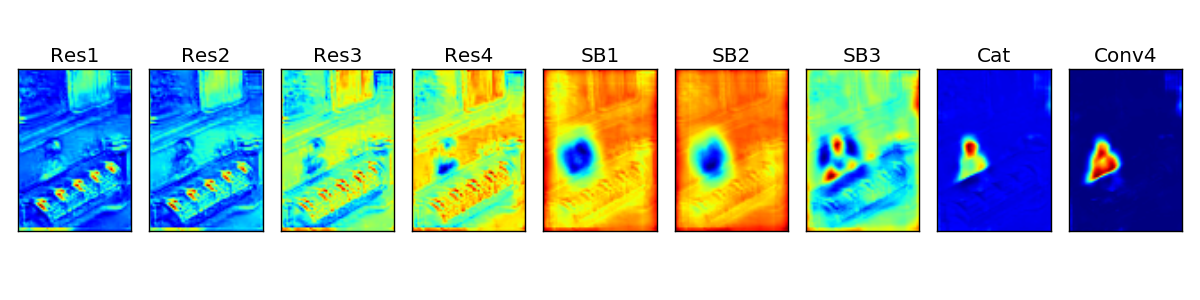}    
\caption{Visualization of activation map extracted from the intermediate layers of ScenePoseNet for different regions of the scene. As the image propagates through ScenePoseNet, the beliefs of the scene background are suppressed while the beliefs on the pedestrian and the individual joints increases.} 
\label{fig:activationmap}
\end{figure*}

For a given specific scene we evaluate the efficacy of our \emph{Visual Compiler} to generate a \emph{Visual Program}, ScenePoseNet, for pedestrian detection, pose estimation and segmentation. Detection and pose estimation are evaluated both quantitatively and qualitatively, while segmentation is evaluated only qualitatively due to lack of ground truth segmentation masks. Figure \ref{fig:activationmap} shows the activation maps at various stages of ScenePoseNet. The spatial belief blocks progressively refine the activation maps from the residual blocks. We note that combining the activation maps from the different spatial belief blocks further improves pedestrian localization in terms of the segmentation mask.

\subsection{Datasets and Baselines}

We evaluated our \emph{Visual Compiler} on two publicly available datasets: (1) \textbf{Towncenter dataset \cite{benfold2011stable}:} This is a video dataset of a semi-crowded town center with a resolution of $1920\times1080$. We down-sample the videos to a standardized resolution of $640\times360$, and (2) \textbf{PETS 2006 dataset \cite{thirde2006overview}:} This datatset consists of video of a train station including a number of pedestrians. From among the four different camera viewpoints in the dataset, we use a single viewpoint for our experiments. We down-sample the videos to a standardized resolution of $640\times512$.

\begin{figure}
    \includegraphics[width=0.48\linewidth, trim={2.1cm 0 3.4cm 0.8cm},clip]{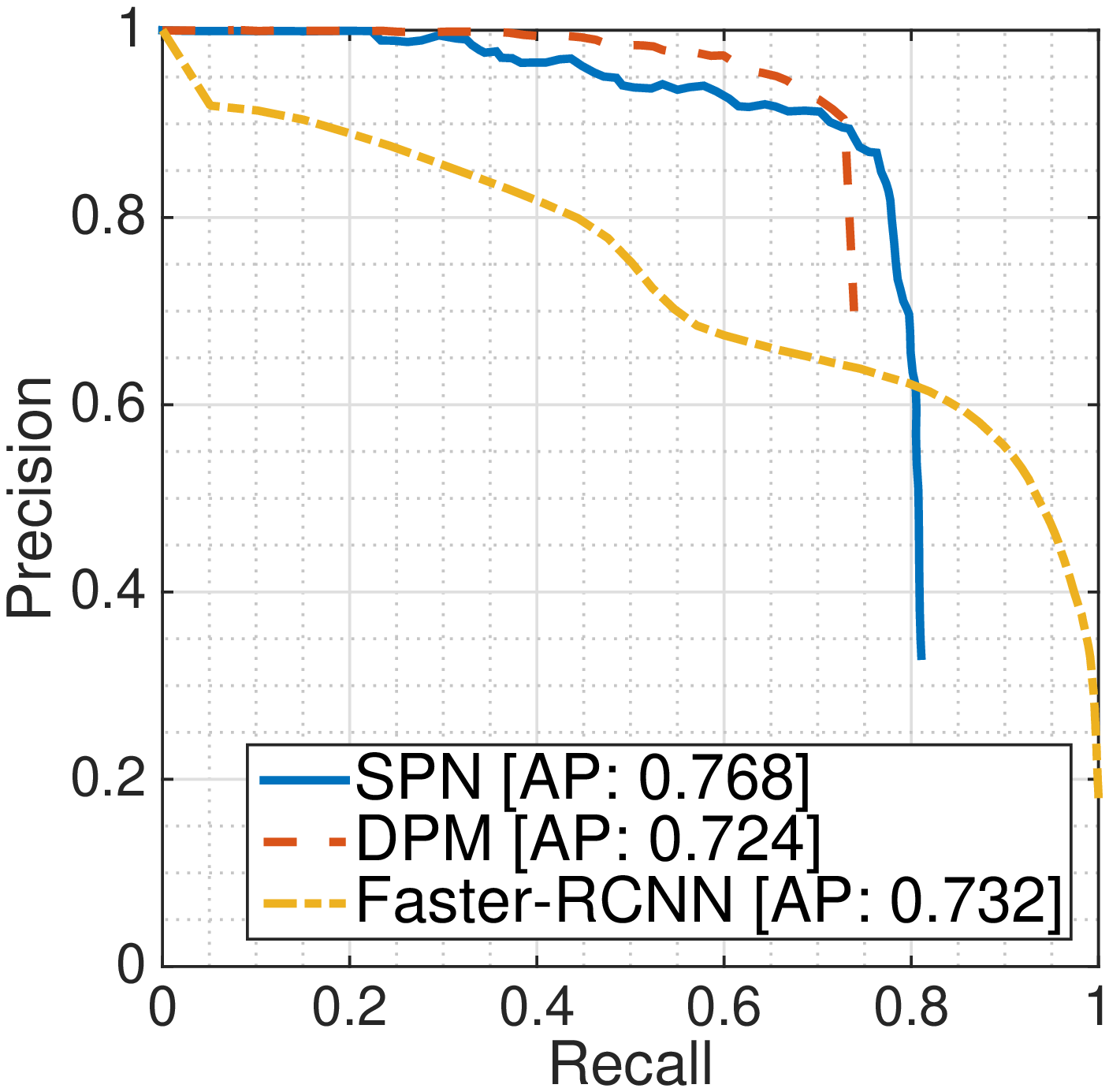}
    \includegraphics[width=0.48\linewidth, trim={2.1cm 0 3.4cm 0.8cm},clip]{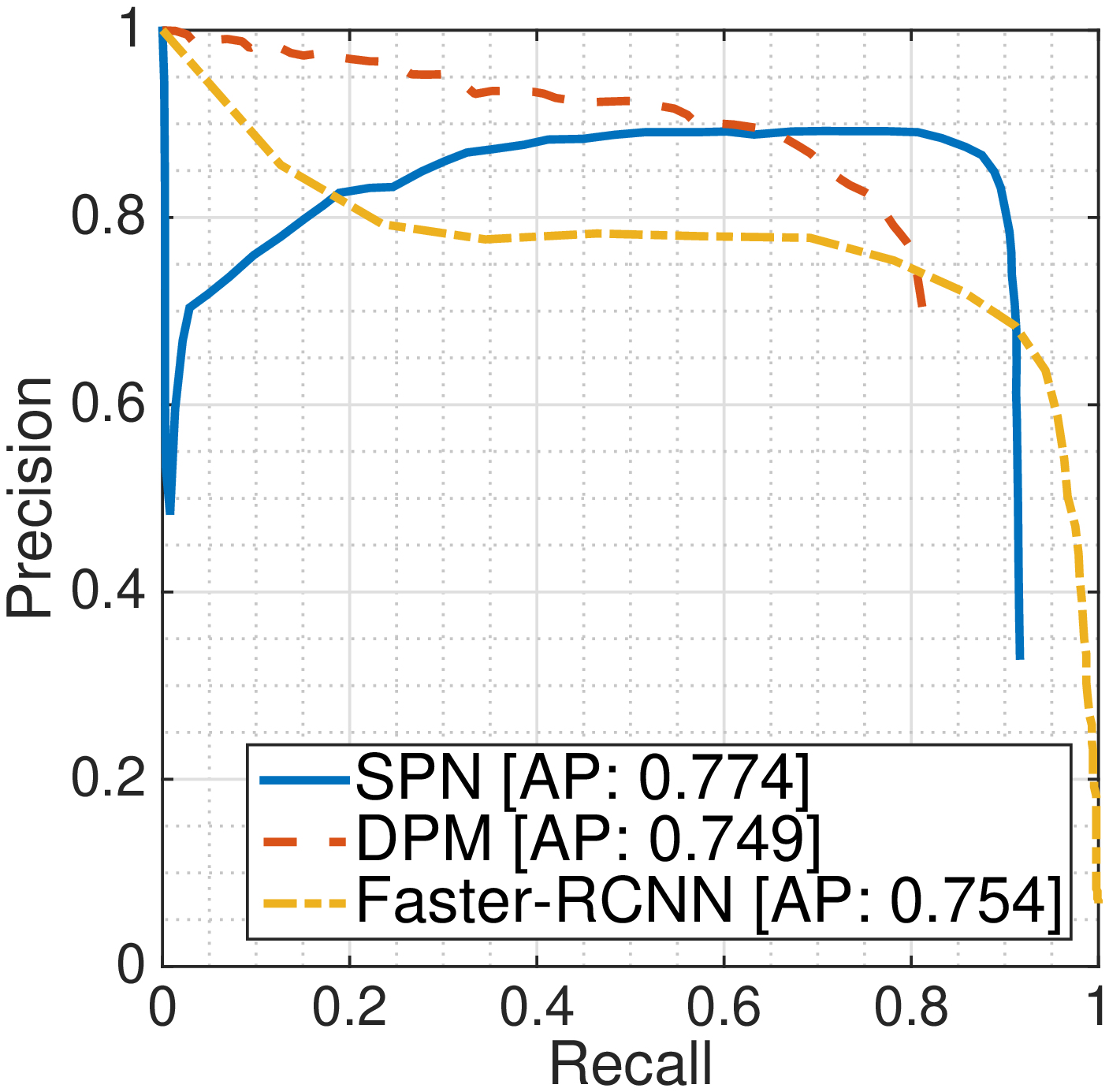}
    \caption{Precision-Recall curves along with the average precision for pedestrian detection on the (a) Towncenter and (b) PETS2006 datasets. We compare our ScenePoseNet (SPN) with DPM and Faster R-CNN.}
    \label{fig:detection_pr_real}
\end{figure}

\begin{figure}[!ht]
	\centering
    \includegraphics[width=0.19\linewidth]{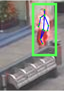}
    \includegraphics[width=0.19\linewidth]{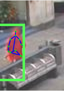}
    \includegraphics[width=0.19\linewidth]{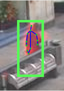}
    \includegraphics[width=0.19\linewidth]{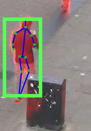}
    \includegraphics[width=0.19\linewidth]{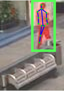}  \\    
    \includegraphics[width=0.19\linewidth]{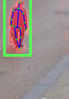}
    \includegraphics[width=0.19\linewidth]{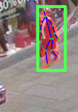}
    \includegraphics[width=0.19\linewidth]{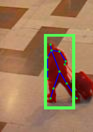}  
    \includegraphics[width=0.19\linewidth]{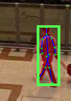} 
    \includegraphics[width=0.19\linewidth]{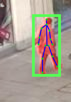} \\
    \includegraphics[width=0.19\linewidth]{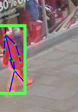}
    \includegraphics[width=0.19\linewidth]{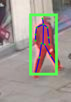} 
    \includegraphics[width=0.19\linewidth]{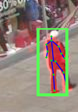}
    \includegraphics[width=0.19\linewidth]{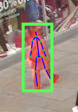}
    \includegraphics[width=0.19\linewidth]{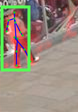}    
    \caption{Qualitative results of our approach predicting \textcolor{green}{bounding box}, body pose in terms of part locations \textcolor{blue}{(skeleton)} and a \textcolor{red}{(segmentation mask)}. The first row shows examples where the pedestrian is occluded.}
\label{fig:qualitativeOutputs}
\end{figure}

We compare our \emph{Visual Compiler} based pedestrian detection and pose estimation approach to a number of combinations of state-of-the-art pedestrian detectors and human pose estimation approaches. For pedestrian detection we consider the two baselines that are based on HoG features, SLSV \cite{hattori2015learning} and Deformable Parts Model (DPM) \cite{yang2013articulated}, and Faster Region-based Convolutional Neural Network \cite{ren2015faster}, pre-trained on ImageNet and VOC2007. For human pose estimation we compare against two state-of-the-art methods, Convolutional Pose Machines (CPM) \cite{wei2016cpm} and Iterative Error Feedback (IEF) \cite{IEF2015human}. Since these methods assume that pedestrians have been detected \emph{a priori}, we use different detectors to localize pedestrians: DPM, Faster R-CNN and varying degrees of jittered ground truth bounding boxes. We also test the ability of CPM and IEF to perform both detection and pose estimation simultaneously as a baseline \emph{i.e.}, using the whole region as the input without localizing the pedestrian.

\subsection{Pedestrian Detection Evaluation}

We compare our ScenePoseNet model to all baselines using the standard $50\%$ overlap metric used for pedestrian detection. Although in theory we can learn a ScenePoseNet model for every location or region in the scene, pedestrians in the datasets tend to walk only in certain parts of the scene. For efficiency, we evaluate detection accuracy on real pedestrians using only high traffic areas. Results are summarized in the precision recall (PR) curve in Figure \ref{fig:detection_pr_real}. The PR curves show that our approach has a significantly better recall rate due to our ability to learn accurate scene-and-region specific detectors.

\begin{table}
\caption{Mean average precision and mean IoU}
\vspace{-0.5cm}
\begin{center}
\begin{tabular}{lcc}
\hline
Method & meanIoU & mAP \\
\hline
Ours & 0.5502 & 0.768 \\
SLSV \cite{hattori2015learning} & 0.4041 & 0.5201 \\
\hline
\end{tabular}
\end{center}
\label{tab:compareCVPR2015}
 \vspace{-5mm}
\end{table}

\begin{figure}[t!]
\centering
    \includegraphics[width=0.8\linewidth, trim={38mm 102mm 130mm 75mm},clip]{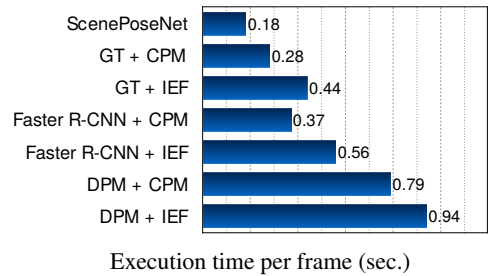}\\
    \small Execution time per frame (sec.)
    \caption{Comparison of speed across different approaches for pose estimation and pedestrian detection.}
    \label{fig:speed}
\end{figure}

Our approach, trained purely on synthetic data, outperforms generic state-of-the-art detectors that are trained on real data. This provides validation for our premise that explicitly making use of scene geometry, obstacles and camera setup can significantly help synthesis based techniques outperform models that are trained on real data. Finally we also compare the performance of our approach with SLSV, which also learns a scene scene-specific pedestrian detection model based on traditional HoG features. 
The comparison of mean average precision by 50\% bounding box overlap and mean IoU at high traffic region in the Towncenter dataset are summarized in Table \ref{tab:compareCVPR2015}. The results show that ScenePoseNet exhibits better localization performance in comparison to SLSV even when both of the approaches leverage scene geometry.

\subsection{Pose Estimation Evaluation}

\begin{figure*}[!ht]
\begin{subfigure}[b]{0.32\linewidth}
    \includegraphics[width=\linewidth, trim={0cm 0cm 0cm -0.4cm}, clip]{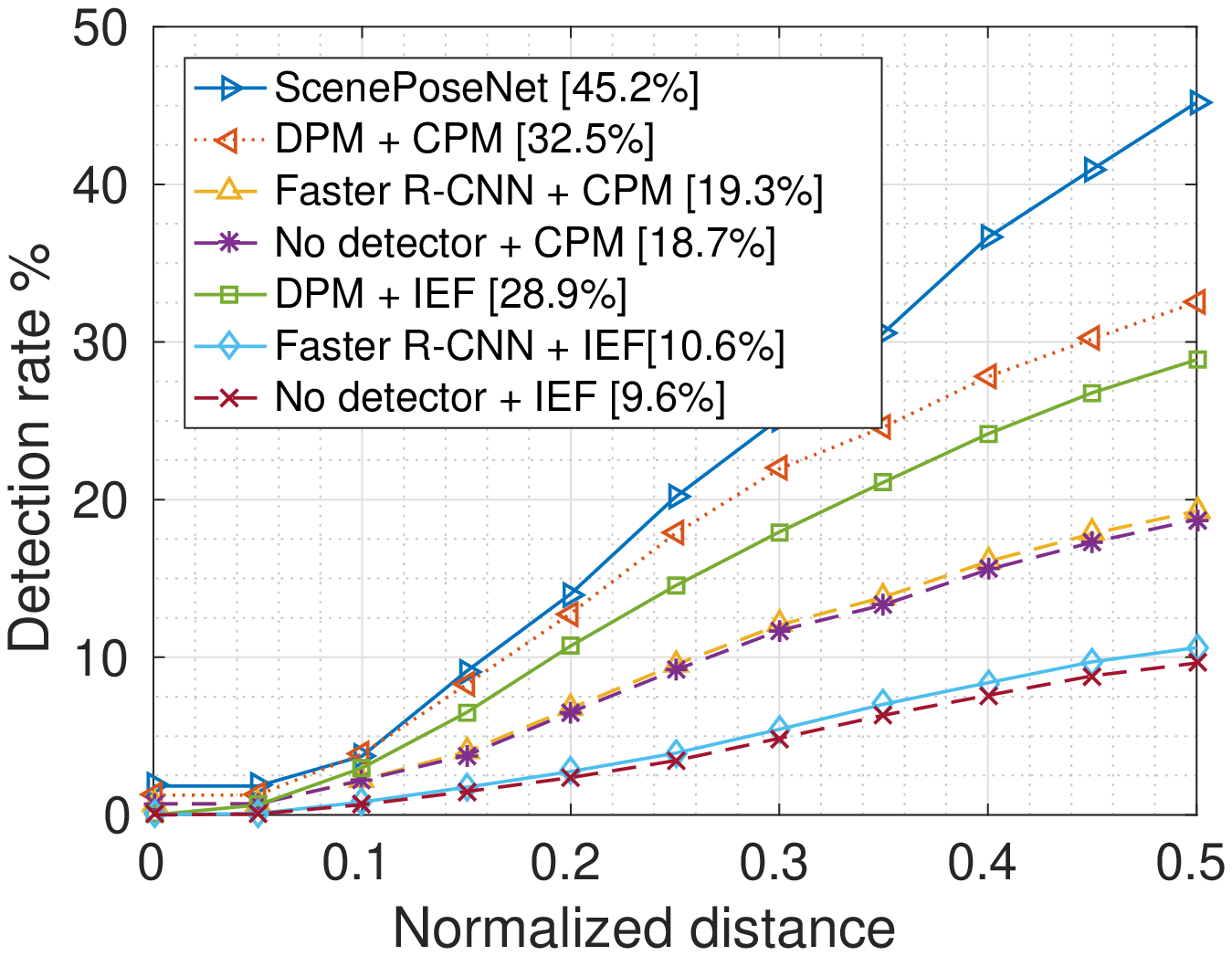}
    \caption{Towncenter}
    \label{fig:pose_towncenter}
\end{subfigure}
\begin{subfigure}[b]{0.32\linewidth}
    \includegraphics[width=\linewidth, trim={0cm 0cm 0cm -0.4cm}, clip]{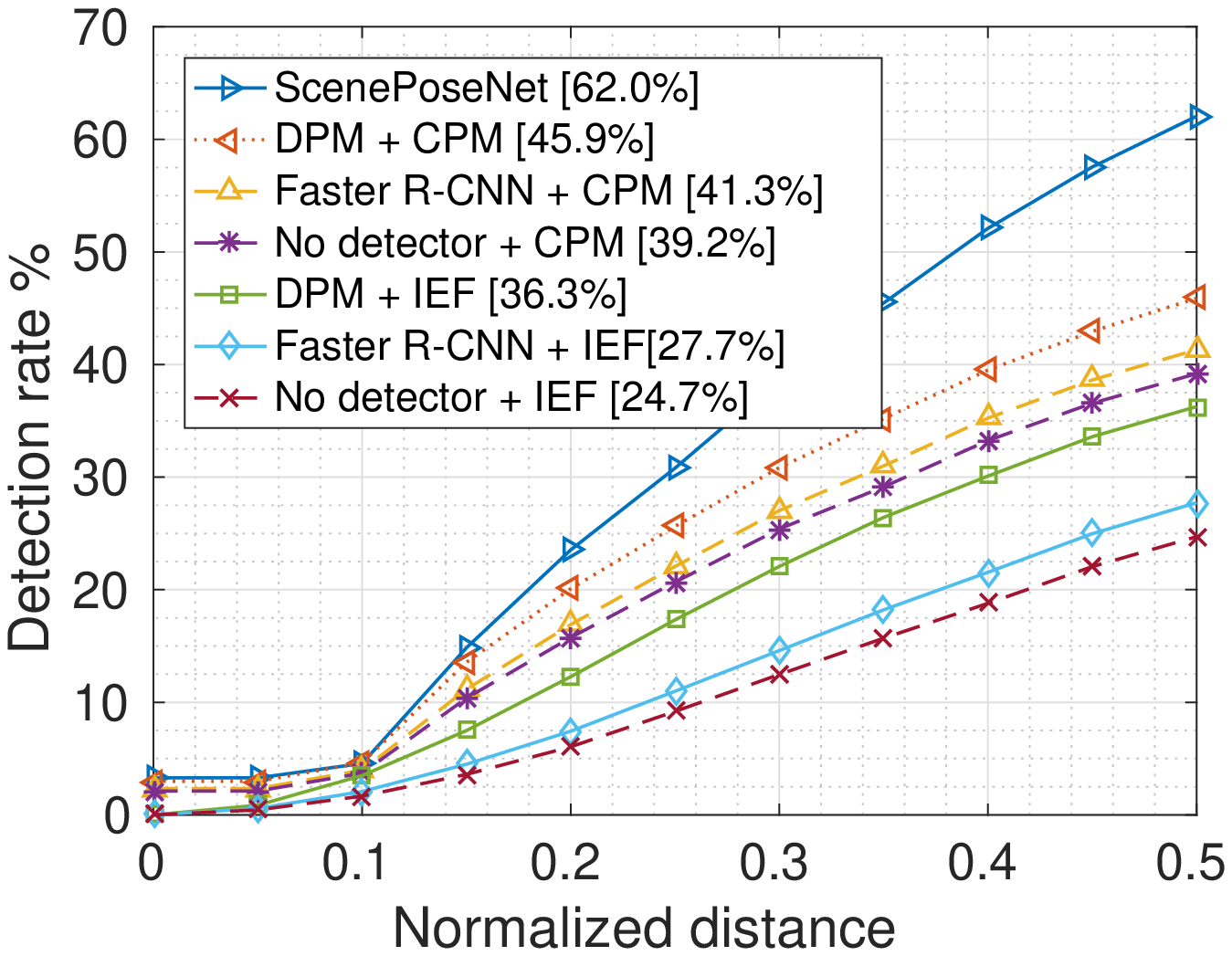}
    \caption{PETS2006}
    \label{fig:pose_pets2006}
\end{subfigure}
\begin{subfigure}[b]{0.32\linewidth}
    \includegraphics[width=\linewidth, trim={0cm 0.1cm 0cm -0.1cm}, clip]{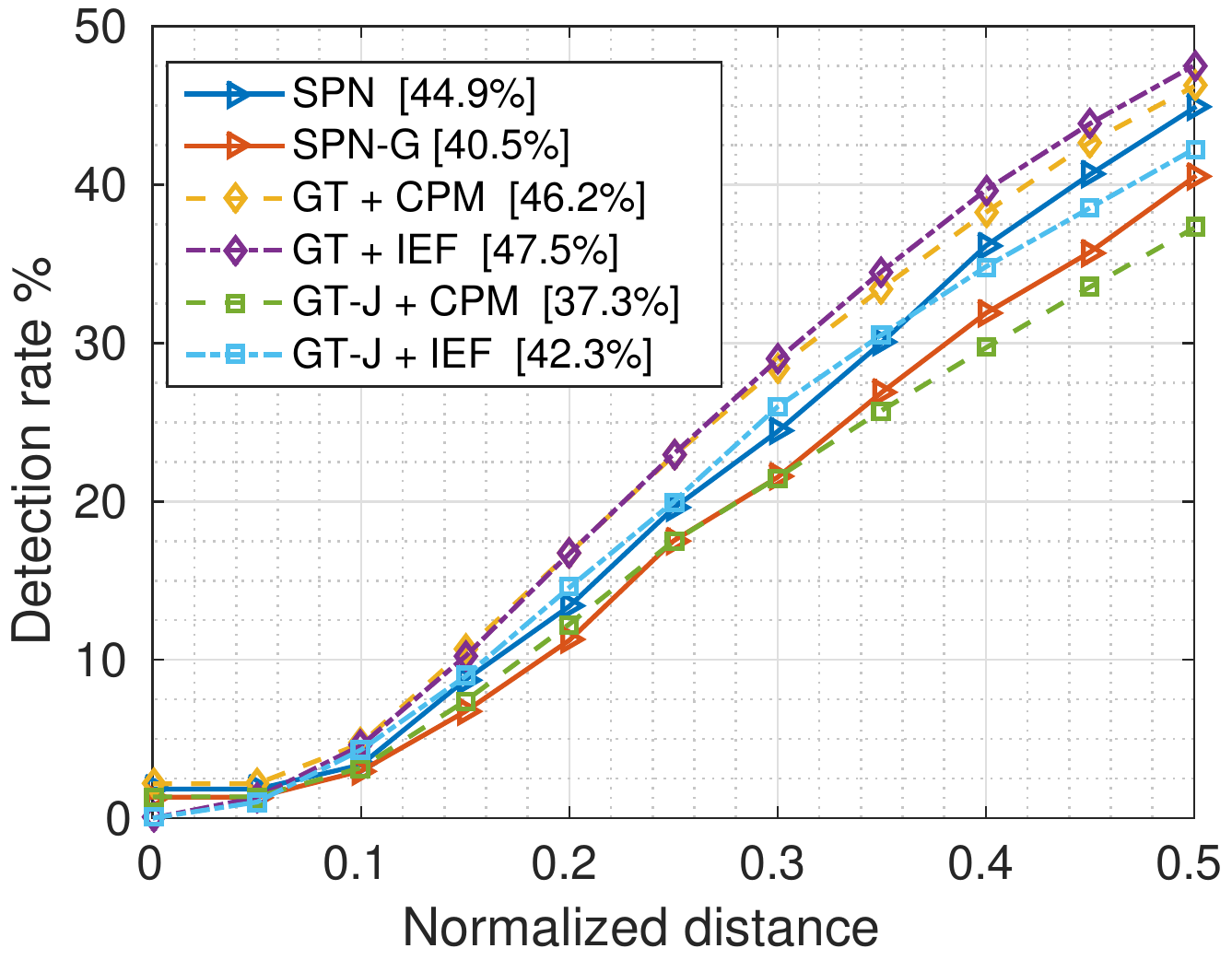}
    \caption{Ground Truth (Towncenter)}
    \label{fig:gt}
\end{subfigure}
\caption{Pose estimation performance on the (a) Towncenter and the (b) PETS2006 dataset against multiple baselines on real data. The number in the bracket corresponds to a PCKh threshold of 0.5. The baselines are combinations of state-of-the-art detection and pose estimation methods as well as pose estimation without pedestrian detection. (c) We also compare pose estimation performance on the Towncenter dataset when using the ground truth (GT) detections and their jittered (GT-J) versions as well as the ScenePoseNet generic-- SPN-G, where we learn a single model for the entire scene.}
\end{figure*}

\begin{figure*}
\begin{subfigure}[b]{0.32\linewidth}
    \includegraphics[width=\linewidth]{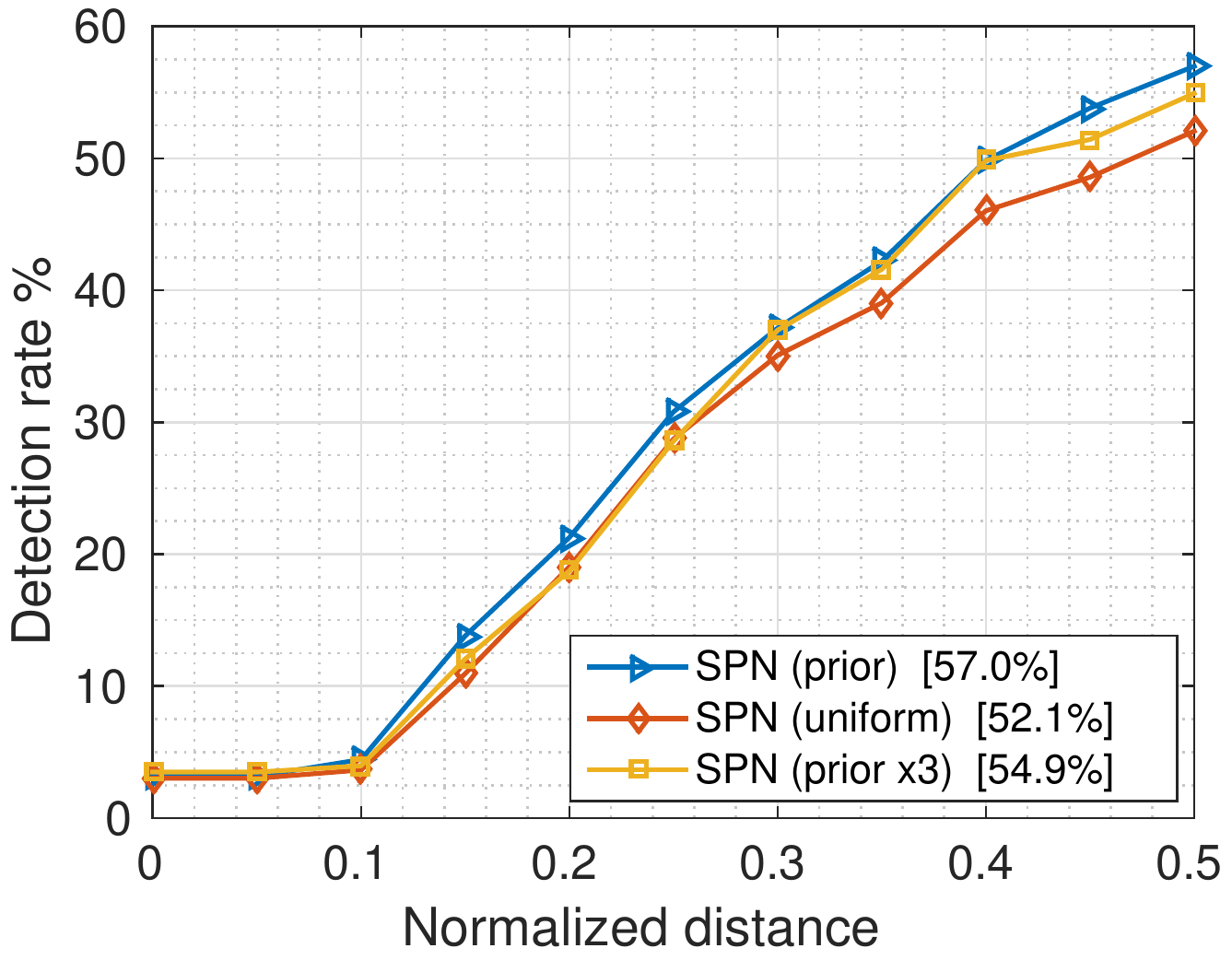}
    \caption{Data Prior}
    \label{fig:prior}
\end{subfigure}
\begin{subfigure}[b]{0.32\linewidth}
    \includegraphics[width=\linewidth]{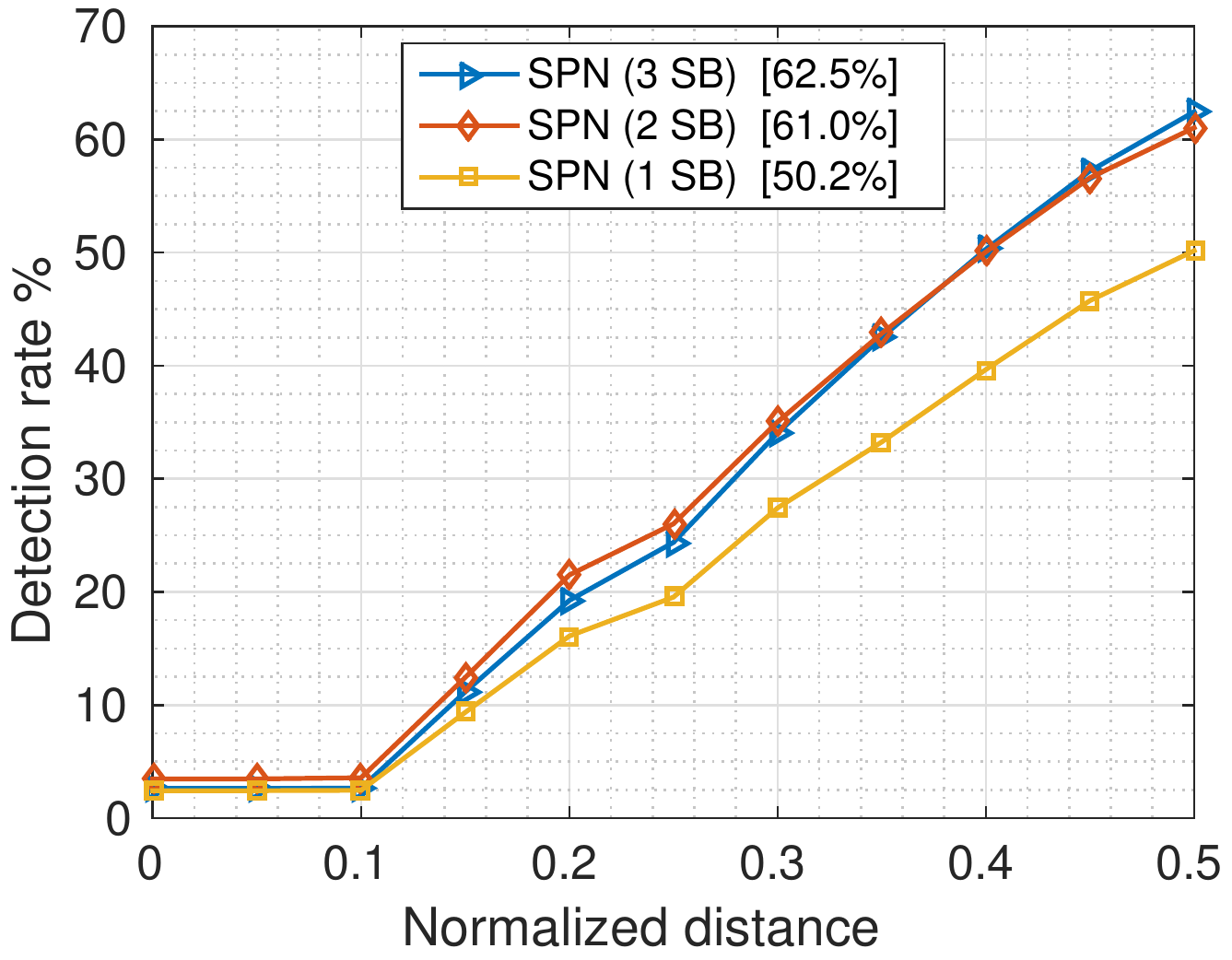}
    \caption{Number of Spatial Belief Units}
    \label{fig:cbunits}
\end{subfigure}
\begin{subfigure}[b]{0.32\linewidth}
    \includegraphics[width=\linewidth]{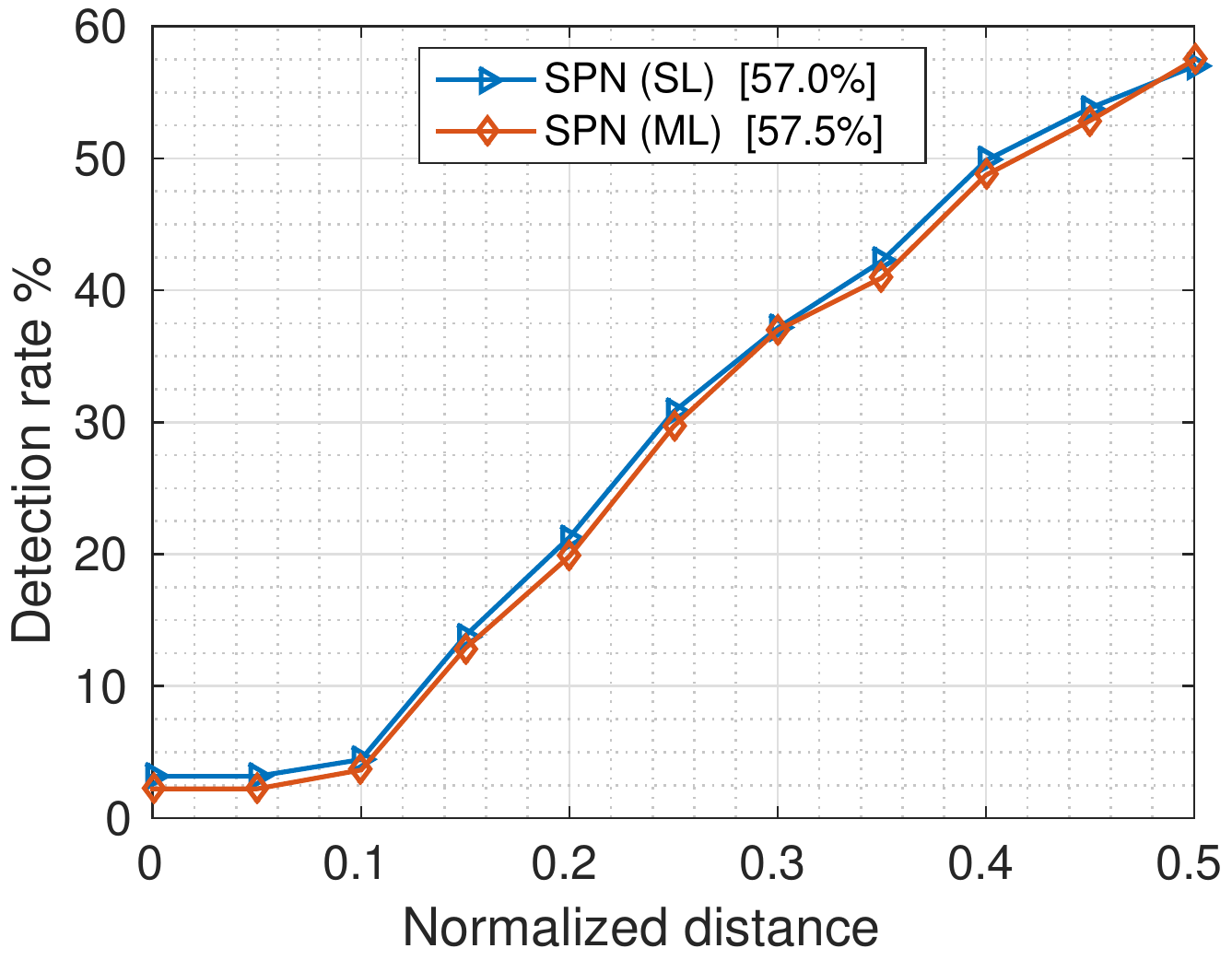}
    \caption{Intermediate Supervision}
    \label{fig:supervision}
\end{subfigure}
\caption{Pose estimation results on real data: (a) here we demonstrate the advantage of using a data prior for sampling pedestrian orientation and pose, (b) exploring the effect of number of spatial belief units, and (c) the effect of training our model with intermediate supervision i.e., optimizing a single loss function (SL) and multiple loss functions (ML).}
\end{figure*}

We compare our ScenePoseNet model to all baselines using the standard PCKh metric used for pose estimation. We evaluate pose estimation accuracy both on synthetically\footnote{Due to lack of space, all results on synthetic data can be found in the supplementary material. Notice that ScenePoseNet outperforms all the baselines on the synthetic images by a large margin.} rendered pedestrians and real pedestrians. Results are summarized as a function of overlap threshold in Figure \ref{fig:pose_towncenter} and Figure \ref{fig:pose_pets2006} for the Towncenter and the PETS2006 datasets respectively. Our approach outperforms all the baselines on real pedestrian data from the two scenes without using any real data for training. By generating physically grounded and geometrically accurate renderings of pedestrians along with high-quality segmentation masks and noise free joint annotations, ScenePoseNet is able to bridge the gap between real and synthetic data. In Figure \ref{fig:qualitativeOutputs}, qualitative results are provided at different regions on both PETS2006 and Towncenter datasets.


Finally, we quantify the performance of just the pose estimator by presenting the baseline pose estimators with ground truth pedestrian detection and randomly jittered ground truth (to simulate a better detector). We also compare against a variant of ScenePoseNet (SPN-G), that learns only one \emph{general} network for the entire scene and is not tuned to any specific region of the scene. Figure \ref{fig:gt} shows this comparison along with the SPN-G variant. The SPN-G variant that is trained for the entire dataset also outperforms the generic pose estimators.


\subsection{Time Complexity}

We compare the inference time complexity of ScenePoseNet and the baselines for detection, pose estimation and the combined task. The timing results are summarized in Figure \ref{fig:speed}. We used code provided by the authors for the baselines and all timing measurements were performed on the same computational setup with an Intel i7-5390 processor with a single Titan-X GPU. The time for the joint task of detection and pose estimation depends on the respective detection and pose estimation baseline combinations. By coupling the tasks of human localization and pose estimation into a single network, ScenePoseNet is significantly, over 100\%, faster than the fastest baseline combination, Faster-RCNN + CPM. ScenePoseNet processes each frame in 0.18sec while the baseline combination takes around 0.37sec for both detection and pose estimation.

\subsection{Ablative Analysis}

\textbf{Effect of data:} Here we study the effect of rendering data with prior knowledge and the amount of data being used. We perform the following comparisons (see Figure \ref{fig:prior}): (1) 50,000 training renders sampled from a prior distribution of pedestrian orientation and pose, (2) 50,000 training renders sampled from a uniform distribution of orientation and pose, and (3) 150,000 training renders sampled from a prior distribution of pedestrian orientation and pose. Leveraging prior knowledge on the likely orientation and pose of people in the scene allows us make effective use of the rendered data. Furthermore, we observed that using more than 50,000 training images does not improve the performance on real images, therefore we use 50,000 images to train our models, sampled from a prior distribution for the Towncenter and from a uniform distribution for the PETS2006 dataset.

\textbf{Effect of SB units:} Here we study the effect of the number of stacked SB units (see Figure \ref{fig:cbunits} for the results). We observe that using two SB gives a significant boost over using a single SB unit. Adding one more SB unit does not seem to help with the synthetic data but provides a slight performance boost on real data.

\textbf{Intermediate supervision and skip connections:} We evaluate the effectiveness of using intermediate supervision, as suggested by some recent pose estimation approaches \cite{wei2016cpm}, on the performance of ScenePoseNet. We train and evaluate our network with intermediate supervision at the outputs of the spatial-belief units. Figure \ref{fig:supervision} shows the comparison. The network using intermediate supervision only provides a small gain in performance. Therefore, we do not use intermediate supervision in our experiments. Finally, we note that the skip connections have proven critical in being able to train our network. Repeated attempts to train our network without the skip connections has resulted in convergence failure. 

\section{Conclusion}
\vspace{-2mm}
We introduced the \emph{Visual Compiler} framework that converts a high level specification of a scene, in terms of camera parameters and other geometric attributes, into ready to use models for object detection and pose estimation. The \emph{Visual Compiler} generates a \emph{Visual Program} in the form of a deep convolutional neural network trained on scene specific synthetic data. The rendering system generates physically grounded and geometrically plausible renders of synthetic humans that serve as training data for our scene-specific pedestrian detection and pose estimation model. Our experimental results suggest a surprising outcome, the \emph{Visual Compiler} can effectively generate a pedestrian detector and pose estimator just from a high level description of the scene. The models compiled by our framework can serve as an alternative to using state-of-the-art off-the-shelf generic for pedestrian detection and pose estimation.


{\small
\bibliographystyle{ieee}
\bibliography{mybib}

\begin{thebibliography}{10}\itemsep=-1pt

\bibitem{aubry2014seeing}
M.~Aubry, D.~Maturana, A.~Efros, B.~Russell, and J.~Sivic.
\newblock Seeing 3d chairs: exemplar part-based 2d-3d alignment using a large
  dataset of cad models.
\newblock In {\em CVPR}, 2014.

\bibitem{benfold2011stable}
B.~Benfold and I.~Reid.
\newblock Stable multi-target tracking in real-time surveillance video.
\newblock In {\em CVPR}, 2011.

\bibitem{cai2015learning}
Z.~Cai, M.~Saberian, and N.~Vasconcelos.
\newblock Learning complexity-aware cascades for deep pedestrian detection.
\newblock In {\em ICCV}, 2015.

\bibitem{IEF2015human}
J.~Carreira, P.~Agrawal, K.~Fragkiadaki, and J.~Malik.
\newblock Human pose estimation with iterative error feedback.
\newblock In {\em CVPR}, 2016.

\bibitem{dalal2005histograms}
N.~Dalal and B.~Triggs.
\newblock Histograms of oriented gradients for human detection.
\newblock In {\em CVPR}, 2005.

\bibitem{dollar2009integral}
P.~Doll{\'a}r, Z.~Tu, P.~Perona, and S.~Belongie.
\newblock Integral channel features.
\newblock In {\em BMVC}, 2009.

\bibitem{felzenszwalb2005pictorial}
P.~Felzenszwalb and D.~Huttenlocher.
\newblock Pictorial structures for object recognition.
\newblock {\em IJCV}, 2005.

\bibitem{fischer2015flownet}
P.~Fischer, A.~Dosovitskiy, E.~Ilg, P.~H{\"a}usser, C.~Haz{\i}rba{\c{s}},
  V.~Golkov, P.~van~der Smagt, D.~Cremers, and T.~Brox.
\newblock Flownet: Learning optical flow with convolutional networks.
\newblock In {\em ICCV}, 2015.

\bibitem{girshick2015fast}
R.~Girshick.
\newblock Fast r-cnn.
\newblock In {\em ICCV}, 2015.

\bibitem{girshick2011object}
R.~Girshick, P.~Felzenszwalb, and D.~Mcallester.
\newblock Object detection with grammar models.
\newblock In {\em NIPS}, 2011.

\bibitem{hattori2015learning}
H.~Hattori, V.~Boddeti, K.~Kitani, and T.~Kanade.
\newblock Learning scene-specific pedestrian detectors without real data.
\newblock In {\em CVPR}, 2015.

\bibitem{he2016identity}
K.~He, X.~Zhang, S.~Ren, and J.~Sun.
\newblock Identity mappings in deep residual networks.
\newblock {\em arXiv preprint arXiv:1603.05027}, 2016.

\bibitem{karpathy2015deep}
A.~Karpathy and L.~Fei-Fei.
\newblock Deep visual-semantic alignments for generating image descriptions.
\newblock In {\em CVPR}, 2015.

\bibitem{krizhevsky2012imagenet}
A.~Krizhevsky, I.~Sutskever, and G.~Hinton.
\newblock Imagenet classification with deep convolutional neural networks.
\newblock In {\em NIPS}, 2012.

\bibitem{wei2016ssd}
W.~Liu, D.~Anguelov, D.~Erhan, and C.~Szegedy.
\newblock {SSD}: Single shot multibox detector.
\newblock In {\em ECCV}, 2016.

\bibitem{long2015fully}
J.~Long, E.~Shelhamer, and T.~Darrell.
\newblock Fully convolutional networks for semantic segmentation.
\newblock In {\em CVPR}, 2015.

\bibitem{newell2016stacked}
A.~Newell, K.~Yang, and J.~Deng.
\newblock Stacked hourglass networks for human pose estimation.
\newblock {\em arXiv preprint arXiv:1603.06937}, 2016.

\bibitem{ouyang2013joint}
W.~Ouyang and X.~Wang.
\newblock Joint deep learning for pedestrian detection.
\newblock In {\em ICCV}, 2013.

\bibitem{pishchulin2013strong}
L.~Pishchulin, M.~Andriluka, P.~Gehler, and B.~Schiele.
\newblock Strong appearance and expressive spatial models for human pose
  estimation.
\newblock In {\em ICCV}, 2013.

\bibitem{pishchulin2012articulated}
L.~Pishchulin, A.~Jain, M.~Andriluka, T.~Thorm{\"a}hlen, and B.~Schiele.
\newblock Articulated people detection and pose estimation: Reshaping the
  future.
\newblock In {\em CVPR}, 2012.

\bibitem{ramakrishna2014pose}
V.~Ramakrishna, D.~Munoz, M.~Hebert, A.~Bagnell, and Y.~Sheikh.
\newblock Pose machines: Articulated pose estimation via inference machines.
\newblock In {\em ECCV}, 2014.

\bibitem{ren2015faster}
S.~Ren, K.~He, R.~Girshick, and J.~Sun.
\newblock Faster r-cnn: Towards real-time object detection with region proposal
  networks.
\newblock In {\em NIPS}, 2015.

\bibitem{shotton2013efficient}
J.~Shotton, R.~Girshick, A.~Fitzgibbon, T.~Sharp, M.~Cook, M.~Finocchio,
  R.~Moore, P.~Kohli, A.~Criminisi, A.~Kipman, et~al.
\newblock Efficient human pose estimation from single depth images.
\newblock {\em PAMI}, 2013.

\bibitem{su2015render}
H.~Su, C.~Qi, Y.~Li, and L.~Guibas.
\newblock Render for cnn: Viewpoint estimation in images using cnns trained
  with rendered 3d model views.
\newblock In {\em ICCV}, 2015.

\bibitem{taigman2014deepface}
Y.~Taigman, M.~Yang, M.~Ranzato, and L.~Wolf.
\newblock Deepface: Closing the gap to human-level performance in face
  verification.
\newblock In {\em CVPR}, 2014.

\bibitem{thirde2006overview}
D.~Thirde, L.~Li, and F.~Ferryman.
\newblock Overview of the pets2006 challenge.
\newblock In {\em IEEE International Workshop on Performance Evaluation of
  Tracking and Surveillance (PETS 2006)}, 2006.

\bibitem{tian2015deep}
Y.~Tian, X.~W. P.~Luo, and X.~Tang.
\newblock Deep learning strong parts for pedestrian detection.
\newblock In {\em ICCV}, 2015.

\bibitem{toshev2014deeppose}
A.~Toshev and C.~Szegedy.
\newblock Deeppose: Human pose estimation via deep neural networks.
\newblock In {\em CVPR}, 2014.

\bibitem{vazquez2014virtual}
D.~Vazquez, A.~Lopez, J.~Marin, D.~Ponsa, and D.~Geronimo.
\newblock Virtual and real world adaptation for pedestrian detection.
\newblock {\em PAMI}, 2014.

\bibitem{wei2016cpm}
S.~Wei, V.~Ramakrishna, T.~Kanade, and Y.~Sheikh.
\newblock Convolutional pose machines.
\newblock In {\em CVPR}, 2016.

\bibitem{wei2016deformablepose}
W.~Yang, W.~Ouyang, H.~Li, and X.~Wang.
\newblock End-to-end learning of deformable mixture of parts and deep
  convolutional neural networks for human pose estimation.
\newblock In {\em CVPR}, 2016.

\bibitem{yang2013articulated}
Y.~Yang and D.~Ramanan.
\newblock Articulated human detection with flexible mixtures of parts.
\newblock {\em PAMI}, 2013.

\bibitem{zhang2015filtered}
S.~Zhang, R.~Benenson, and B.~Schiele.
\newblock Filtered channel features for pedestrian detection.
\newblock In {\em CVPR}, 2015.

\end{thebibliography}
}

\end{document}